\documentclass[sigconf]{acmart}

\AtBeginDocument{%
  }

\copyrightyear{2025}
\acmYear{2025}
\setcopyright{rightsretained}
\acmConference[EICS Companion '25]{Companion Proceedings of the 17th ACM SIGCHI Symposium on Engineering Interactive Computing Systems}{June 23--27, 2025}{Trier, Germany}
\acmBooktitle{Companion Proceedings of the 17th ACM SIGCHI Symposium on Engineering Interactive Computing Systems (EICS Companion '25), June 23--27, 2025, Trier, Germany}\acmDOI{10.1145/3731406.3731970}
\acmISBN{979-8-4007-1866-3/2025/06}

\begin{document}

\title[CBM-RAG: Demonstrating Enhanced Interpretability...]{CBM-RAG: Demonstrating Enhanced Interpretability in Radiology Report Generation with Multi-Agent RAG and Concept Bottleneck Models}

\author{Hasan Md Tusfiqur Alam}
\email{hasan.alam@dfki.de}
\orcid{0000-0003-1479-7690}
\affiliation{
    \institution{German Research Center for Artificial Intelligence (DFKI)}
    \city{Saarbrücken}
    \country{Germany}
}

\author{Devansh Srivastav}
\email{devansh.srivastav@dfki.de}
\orcid{0000-0002-8858-7402}
\affiliation{
    \institution{German Research Center for Artificial Intelligence (DFKI)}
    \city{Saarbrücken}
    \country{Germany}
}
\affiliation{
    \institution{Saarland University}
    \city{Saarbrücken}
    \country{Germany}
}

\author{Abdulrahman Mohamed Selim}
\email{abdulrahman.mohamed@dfki.de}
\orcid{0000-0002-4984-6686}
\affiliation{
    \institution{German Research Center for Artificial Intelligence (DFKI)}
    \city{Saarbrücken}
    \country{Germany}
}

\author{Md Abdul Kadir}
\orcid{0000-0002-8420-2536}
\email{abdul.kadir@dfki.de}
\affiliation{
    \institution{German Research Center for Artificial Intelligence (DFKI)}
    \city{Saarbrücken}
    \country{Germany}
}
\affiliation{
    \institution{University of Oldenburg}
    \city{Oldenburg}
    \country{Germany}
}

\author{Md Moktadirul Hoque Shuvo}
\orcid{0009-0009-0888-7906}
\affiliation{
    \institution{Dhaka Medical College Hospital}
    \city{Dhaka}
    \country{Bangladesh}
}

\author{Daniel Sonntag}
\orcid{0000-0002-8857-8709}
\email{daniel.sonntag@dfki.de}
\affiliation{
    \institution{German Research Center for Artificial Intelligence (DFKI)}
    \city{Saarbrücken}
    \country{Germany}
}
\affiliation{
    \institution{University of Oldenburg}
    \city{Oldenburg}
    \country{Germany}
}

\renewcommand{\shortauthors}{Alam, et al.}

\begin{abstract}
  Advancements in generative Artificial Intelligence (AI) hold great promise for automating radiology workflows, yet challenges in interpretability and reliability hinder clinical adoption. This paper presents an automated radiology report generation framework that combines Concept Bottleneck Models (CBMs) with a Multi-Agent Retrieval-Augmented Generation (RAG) system to bridge AI performance with clinical explainability. CBMs map chest X-ray features to human-understandable clinical concepts, enabling transparent disease classification. Meanwhile, the RAG system integrates multi-agent collaboration and external knowledge to produce contextually rich, evidence-based reports. Our demonstration showcases the system’s ability to deliver interpretable predictions, mitigate hallucinations, and generate high-quality, tailored reports with an interactive interface addressing accuracy, trust, and usability challenges. This framework provides a pathway to improving diagnostic consistency and empowering radiologists with actionable insights. 
\end{abstract}

\ccsdesc[500]{Applied computing~Health informatics}
\ccsdesc[500]{Computing methodologies~Information extraction}
\ccsdesc[500]{Computing methodologies~Multi-agent systems}
\ccsdesc[300]{Human-centered computing~Heat maps}
\ccsdesc[300]{Information systems~Multimedia and multimodal retrieval}

\keywords{Interpretable Radiology report generation, Disease classification, Medical imaging, Concept Bottleneck Models (CBM), Retrieval-Augmented Generation (RAG), Information Retrieval, VLMs, LLMs.}

\maketitle

\section{Introduction}

Recent advancements in generative models have accelerated computer-aided interpretation for chest X-ray (CXR) images \cite{chen2024chexagent,wang2022medclipcontrastivelearningunpaired,zhang2024biomedclipmultimodalbiomedicalfoundation}. These end-to-end architectures not only predict specific findings but also generate comprehensive radiological reports by integrating a language module \cite{liu2023tailoring,wang2023r2gengpt}. A system that can classify diseases from CXR images and produce coherent reports can reduce radiologists' workload and improve diagnostic consistency. 
However, since large language models (LLMs) are prone to hallucinations \cite{perkovic2024hallucinations}, such generators face reliability issues. 
To address similar challenges in other domains, researchers have introduced Retrieval-Augmented Generation (RAG) \cite{lewis2020retrieval}, which leverages external resources to produce more accurate and reliable conclusions. 
However, the black-box nature of LLMs remains a significant limitation \cite{luo2024understanding}, as they fail to provide explanations or interpretable relationships between inputs and outputs, leading to a system that may be perceived as unreliable and untrustworthy.
Trust in these systems requires transparency \cite{eke2024role}, interpretability \cite{kathait2024comprehensive}, and integration of additional data such as patient history and recent research.

To address these challenges, we propose a conversational tool integrating Concept Bottleneck Models (CBMs) \cite{koh2020concept} with a multi-agent RAG framework to enhance accuracy, interpretability, and reliability in CXR report generation. 
CBMs map visual features to human-understandable clinical concepts and use saliency techniques to highlight relevant image regions, while RAG dynamically incorporates external knowledge, including patient history, prior studies, and current research, to produce evidence-based reports. 
In this paper, we demonstrate an end-to-end implementation that combines interpretable disease classification with robust report generation, mitigating issues of hallucination and opacity, and thereby enhancing AI-driven CXR interpretation for clinical practice to empower radiologists with actionable insights to improve their diagnostic consistency and trust in our system.
Our code\footnote{Code: \url{https://github.com/tifat58/enhanced-interpretable-report-generation-demo.git}} and demo\footnote{Online Demo: \url{https://cxr-cbm-rag-dfki-iml-demo.streamlit.app/}} are publicly available online.

\section{Methodology}

\begin{figure}[t!]
\centering
  \includegraphics[width=\linewidth]{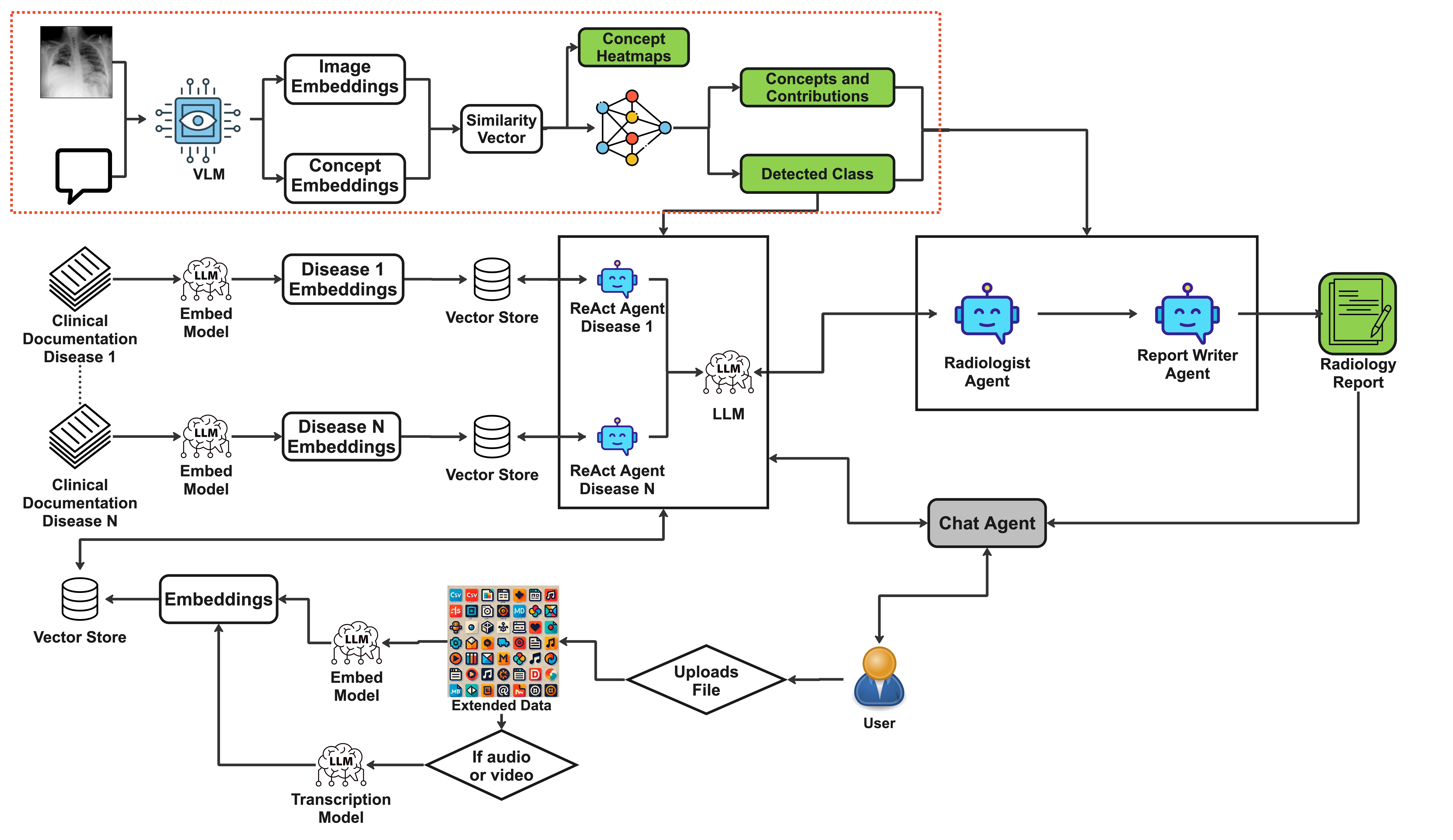}
  \caption{Workflow of the CBM-RAG Framework for Radiology Report Generation. The upper section processes chest X-rays via a VLM to generate clinical concepts, heatmaps, and contribution scores. The lower section uses multi-agent RAG. A Radiologist Agent synthesizes findings, a Report Writer Agent creates detailed reports, and a Chat Agent enables real-time interaction.}
  \Description{This diagram illustrates the complete workflow of the CBM-RAG framework, integrating Concept Bottleneck Models (CBMs) and a multi-agent Retrieval-Augmented Generation (RAG) system. The upper section highlights the interpretable classification pipeline, where a Vision-Language Model (VLM) generates image embeddings for the chest X-ray. These embeddings are compared with concept embeddings to form a similarity vector. Saliency heatmaps for identified concepts are generated, providing visual explanations of detected findings. The detected class and associated concepts are then fed into the multi-agent RAG system. In the lower section, the multi-agent RAG architecture incorporates external knowledge sources. Disease-specific embeddings are retrieved from vector stores populated with clinical documentation, while user-uploaded files, such as audio, video, or text, are transcribed and embedded for retrieval. Dedicated agents for disease-specific analysis (ReAct Agents), radiology interpretation (Radiologist Agent), and final report synthesis (Report Writer Agent) collaborate to produce a comprehensive radiology report. A Chat Agent enables user interaction, facilitating queries and contextual updates throughout the process. This framework emphasizes transparency, interpretability, and adaptability in radiology report generation.}
  \label{fig:flow}
\end{figure}

Our approach starts with a concept bottleneck mechanism \cite{koh2020concept} to identify and quantify medically relevant concepts in a CXR image. Building on prior works \cite{alam2024towards, yan2023robust, oikarinen2023label}, we use LLMs to automatically acquire a set of concepts for classification, rather than relying on manual identification. As shown in Fig. \ref{fig:flow}, we obtain image embedding and text embeddings for the uploaded image and concept set, respectively, from ChexAgent \cite{chen2024chexagent}, a VLM fine-tuned for CXR interpretation, and the Mistral embed model\cite{jiang2023mistral}. We calculate cosine similarity between image embeddings and each text embedding in the concept set to form a similarity matrix. To focus on the most significant features, max pooling is applied to the similarity matrix to form a concept vector. This concept vector is normalized to a scale between $0$ and $1$ for interpretability and fed into a fully connected layer. A classification model uses this vector as input to predict the disease class. We used the COVID-QU dataset \cite{chowdhury2020can}, comprising $33,920$ CXR images with three classes: Pneumonia, COVID-19, and Normal. Finally, the cross-product of the model's weight matrix and the concept vector provides contribution scores, quantifying the influence of each concept on the classification decision. Saliency regions for each concept in the image are derived from the similarity matrix. These heatmaps serve as direct visual indications of how the system localizes concepts such as “pulmonary consolidation” or “nodule” within the CXR image, thereby offering a clear route to interpretability. 


In addition, we used a multi-agent RAG with five specialized agents for report generation. The Pneumonia, COVID-19, and Normal Agents are implemented as Reasoning and Acting (ReAct) agents \cite{yao2022react}. Additionally, the Radiologist Agent interprets clinical concepts using the ReAct agents and queries a pre-configured database from the National Institutes of Health (NIH), while the Report Writer Agent synthesizes the final report. The system also accepts user-provided files (e.g., PDFs, PPTs, text, MP3, MP4), with media transcribed via OpenAI’s Whisper model \cite{radford2023robust} and embedded and indexed for retrieval. This integration enriches reports with updated clinical guidelines, patient histories, and multimedia sources. The framework is implemented using CrewAI and LlamaIndex for efficient retrieval and high-quality report generation.

\section{User Interface}

The user interface (UI) for the CXR analysis system comprises three components: concept generation, report generation, and a conversational chat interface. Upon uploading a CXR image, the concept generation module identifies relevant clinical concepts, computes contribution scores, and predicts disease classes using the CBM. Identified concepts are displayed in an editable list sorted by the absolute values of their contribution scores, each with a toggle for visualizing associated saliency heatmaps. Users can adjust scores to refine model predictions, thereby linking outputs to clinically meaningful features.
After finalizing concept scores, users can generate a comprehensive radiology report. The report generation module integrates clinical documents from trusted sources (e.g., NIH) and accepts additional inputs (text, audio, video, images). The generated report details findings, diagnosis, and guidelines, and an optional chain-of-thought dropdown reveals the multi-agent RAG’s sequential reasoning. A conversational chat interface further enables real-time, context-aware queries regarding the CXR image, report details, or clinical conditions.

\section{Conclusion and Future Work}

In this paper, we presented a tool that bridges AI performance with clinical explainability by linking visual features to human-understandable clinical concepts and integrating external knowledge for context-rich, evidence-based radiology reports. Our framework produces transparent disease classifications and tailored reports while mitigating hallucination and opacity issues. Its interactive UI—with explainable outputs and conversational capabilities—facilitates dynamic clinician engagement, enhancing trust in AI-assisted decision-making. Although technically promising, formal usability studies in real clinical settings 
are yet to be conducted. Future work will include comprehensive user evaluations, extension to other imaging modalities, and exploration of broader healthcare applications.

\begin{acks}
This work was funded by the German Federal Ministry of Education and Research (BMBF) under grant number 01IW23002 (No-IDLE) and by the Endowed Chair of Applied AI at the University of Oldenburg.
\end{acks}

\bibliographystyle{ACM-Reference-Format}
\bibliography{references}

\end{document}